\documentclass[letterpaper, 10 pt, conference]{ieeeconf}  %
\IEEEoverridecommandlockouts
 \usepackage{graphicx}
 \usepackage{subcaption}
 \usepackage{xcolor}
 \usepackage{url}

\title{\LARGE \bf
Play-Testing \textit{REMind}: Evaluating an Educational Robot-Mediated Role-Play Game}

\author{Elaheh Sanoubari$^{1}$, Neil Fernandes$^{2}$, Keith Rebello$^{2}$, Alicia Pan$^{2}$, Andrew Houston$^{3}$, and Kerstin Dautenhahn$^{2}$
\thanks{This research was funded, in part, by the Canada Research Chair Programme in Socially Intelligent Robotics, and the NSERC Canada Graduate Scholarships -- Doctoral (CGS D) program.}
\thanks{$^{1}$Elaheh Sanoubari is with the Department of Systems Design Engineering, University of Waterloo, Canada
        {\tt\small esanouba@uwaterloo.ca}}%
\thanks{$^{2}$Neil Fernandes, Keith Rebello, Alicia Pan, and Kerstin Dautenhahn are with the Department of Electrical \& Computer Engineering, University of Waterloo, Canada
        {\tt\small \{n24ferna, k2rebell, a29pan, kdautenh\}@uwaterloo.ca}}%
\thanks{$^{3}$Andrew Houston is with the Department of Communication Arts, University of Waterloo, Canada
        {\tt\small houston@uwaterloo.ca}}%
}

\begin{document}

\maketitle
\thispagestyle{empty}
\pagestyle{empty}

\begin{abstract}
This paper presents \textit{REMind}, an innovative educational robot-mediated role-play game designed to support anti-bullying bystander intervention among children. \textit{REMind} invites players to observe a bullying scenario enacted by social robots, reflect on the perspectives of the characters, and rehearse defending strategies by puppeteering a robotic avatar. We evaluated \textit{REMind} through a mixed-methods play-testing study with 18 children aged 9--10. The findings suggest that the experience supported key learning goals related to self-efficacy, perspective-taking, understanding outcomes of defending, and intervention strategies. These results highlight the promise of Robot-Mediated Applied Drama (RMAD) as a novel pedagogical framework to support Social-Emotional Learning.
\end{abstract}

\section{INTRODUCTION}
Peer bullying is a persistent problem in schools, with harmful consequences not only for victims, but also for the wider peer group witnessing it. Although many children disapprove of bullying, they often remain passive as bystanders. This inaction is not simply a matter of apathy; rather, it is often shaped by social and emotional barriers such as fear of retaliation, uncertainty about what to do, low confidence, or concern about peer status~\cite{salmivalli2005anti}. Because bystander intervention can stop bullying in many cases, helping children move from silent disapproval to active defending is a critical challenge for anti-bullying interventions.
Traditional anti-bullying instruction often relies on explanation: telling children why bullying is harmful, what the right thing to do is, or what strategies they should ideally use. However, knowing what one should do does not necessarily prepare a child to act in an emotionally charged moment. Supporting bystander intervention requires opportunities for embodied, reflective, and situated practice, where children can rehearse possible strategies, consider consequences, and build confidence.

\begin{figure}[!ht]
  \centering
  \framebox{%
    \includegraphics[
      width=0.9\linewidth,
      clip
    ]{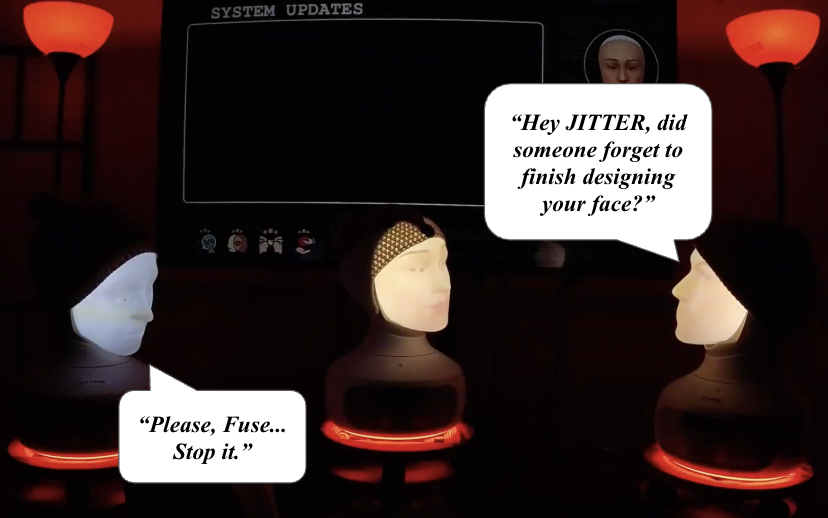}%
  }

  \captionsetup{aboveskip=2pt, belowskip=2pt}
   \caption{\textit{REMind} engages players in role-playing with three social robots to practice anti-bullying bystander intervention.}
    \label{fig:REMindScene}
\end{figure}

Role-play situates learning in experience and allows learners to engage with emotionally and socially complex situations through active participation, which makes it a natural vehicle for developing skills~\cite{blatner2009role}. Hence, it is especially well-suited to address the problem of bystander inaction in peer bullying.
In this paper, we present \textit{REMind\footnote{Short for \textbf{R}obots \textbf{E}mpowering \textbf{Mind}s.}}, an educational robot-mediated role-play game designed to support anti-bullying bystander intervention. \textit{REMind} invites children to observe a bullying scenario enacted by social robots (see Fig. \ref{fig:REMindScene}), reflect on the perspectives of the characters, and rehearse intervention by puppeteering a robotic avatar through their own recorded voice and facial performance. In this way, we use robots not as tutors, but as embodied dramatic proxies through which children can simulate social situations, reflect on emotional nuances, and practice prosocial action.

We report a play-testing study with 18 children aged 9--10 that evaluated whether \textit{REMind} supported key learning goals related to empathy, perspective-taking, outcome expectations, defending strategies, and self-efficacy for defending peers. Our findings combine pre/post quantitative measures with qualitative analysis of children's in-situ reflections during gameplay. The results show that \textit{REMind} significantly increased perceived self-efficacy for defending, shifted children's expectations about the likely outcomes of intervention in theoretically meaningful ways, and elicited rich perspective-taking about the victim, bully, and bystander. Taken together, these findings suggest that Robot-Mediated Applied Drama (RMAD) is a promising pedagogical framework for supporting anti-bullying social-emotional learning.

This paper makes three contributions. First, it presents \textit{REMind}, a novel educational robot-mediated role-play game that uses social robots as embodied dramatic proxies for rehearsing anti-bullying bystander intervention. Second, it reports a mixed-methods play-testing study evaluating whether the experience supported a set of intended learning goals. Third, it provides empirical evidence that robot-mediated role-play can support anti-bullying social-emotional learning by fostering reflection, rehearsal, and perspective-taking in emotionally charged social situations. 
The remainder of the paper reviews related work, describes the design and implementation of \textit{REMind}, presents the study methodology and findings, and concludes with a discussion of limitations, and future directions for real-world deployment at scale.

\section{RELATED WORK}
Social robots are increasingly popular in education, but many applications still cast them primarily as tutor-like agents that deliver curriculum content through explanation-based didactic pedagogy \cite{sanoubari2026aesthetics, belpaeme2018social}. Didactic pedagogy often frames the learner as a passive recipient of knowledge, rather than an active participant in meaning-making—an approach that has been criticized by philosophers such as Jacques Ranci\`ere. In The Ignorant Schoolmaster \cite{ranciere1991ignorant}, Ranci\`ere radically rejects explanation, arguing that it \textit{``stultifies''} learners by reinforcing their dependence on teachers.
He poses that learners achieve \textit{``intellectual emancipation''} when they can use their own resources to solve problems; 
That is, when they are trusted with agency to think, and act for themselves.

This principle resonates with experiential learning approaches like role-play, which invite learners to actively engage with real-world scenarios through embodied participation. Role-play engages all three domains of learning described in Bloom’s Taxonomy: cognitive domain (mental skills, knowledge), affective domain (emotional responses and attitude), and the psychomotor domain (physical skills and abilities)~\cite{bloom1956taxonomy}, which make it a versatile pedagogical tool, widely used in education, therapy, and professional training to help individuals rehearse emotionally or socially complex situations \cite{winardy2023role, daniau2016transformative}. Literature emphasizes the capacity of role-playing for developing empathy, self-awareness, and critical ethical reasoning as the strength of the pedagogical form \cite{bowman2014educational}.  

Role-Playing Games (RPGs) are interactive, story-driven experiences where players assume fictional roles to shape a narrative. Proven to increase learner engagement, they are used in diverse educational subjects \cite{chiu2017role, tarng2010design} and span multiple formats, from digital games to live-action role-playing (LARP) \cite{prager2019exploring, bowman2014educational}. Their primary educational strength lies in providing a safe, simulated environment where learners can explore decisions and their consequences without real-world risk. This protective frame fosters critical thinking and problem-solving skills \cite{winardy2023role, prager2019exploring}, and can even facilitate transformative learning by challenging a person's deeply held assumptions \cite{winardy2023role, daniau2016transformative}.
Central to this process is the mechanism of role-taking, which requires active player agency to make narrative-shaping decisions within the game's fictional reality \cite{bowman2018psychology, walton1993mimesis}. The act of narrative enactment through a character is strongly linked to perspective-taking, a skill that directly cultivates empathy, altruism, and a deeper emotional understanding of others \cite{hammer2018learning, bowman2014educational}.

Role-playing games with virtual agents has been successfully used in anti-bullying and educational interventions. For example, the FearNot! system immerses children in a virtual school where they advise animated characters, a method shown to increase bystander intervention \cite{hall2009fearnot, herkama2018kiva}. Similarly, the KiVa program's game component allows students to role-play as bystanders in realistic scenarios, with large-scale trials demonstrating significant reductions in bullying and victimization \cite{herkama2018kiva, garandeau2023effects}.
However, comparable uses of physically embodied social robots remain underexplored as a pedagogical format, despite the potential of embodiment to heighten social presence and engagement. This is evidenced by related work reporting that compared to virtual agents, robots can elicit more focused attention from children; and consequently, learners consistently demonstrate higher trust, enjoyment and compliance when interacting with them \cite{breazeal2016social}, and they are more adept at building rapport, facilitating therapeutic outcomes, and yielding substantial changes in behaviors and attitudes~\cite{robaczewski2021socially}.
Grounded in this literature, \textit{REMind} leverages the unique power of RPGs and the unique affordances of social robots in order to empower bystanders 
to defend victims of peer bullying.

\section{SYSTEM OVERVIEW}

\begin{figure}[!ht]
  \centering
  \framebox{%
    \includegraphics[
      width=0.95\linewidth,
      clip
    ]{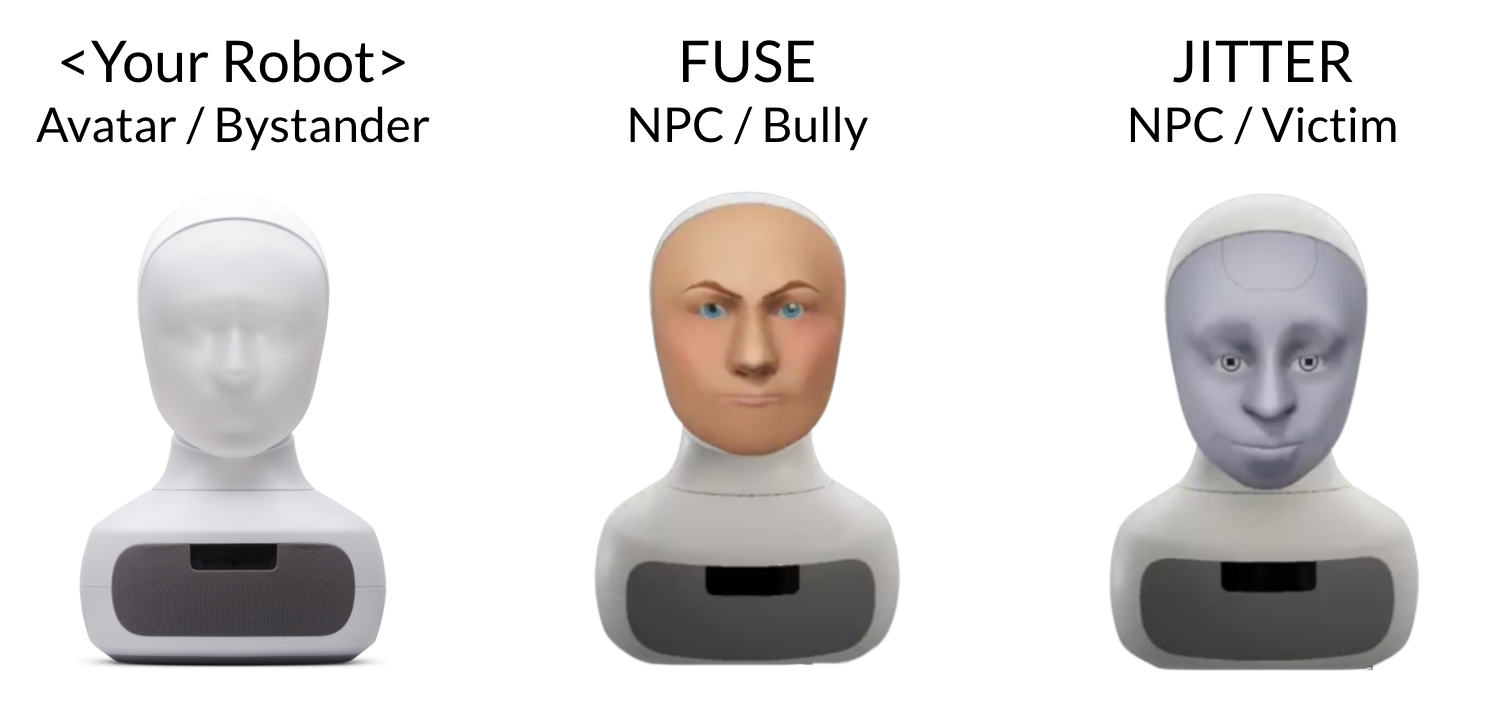}%
  }

  \captionsetup{aboveskip=2pt, belowskip=1pt}
   \caption{Three Robot Characters in REMind.}
    \label{fig:REMindCharacters}
\end{figure}

Using the Robot-Mediated Applied Drama (RMAD)~\cite{sanoubari2022robot-mediated} framework, the system invites players engage in a role-play with three social robots by playing the role of a mentor to a robotic avatar and navigating through a five-phase interactive narrative. The robots play distinct narrative roles: \textit{AVATAR}, is the player’s customizable in-game avatar; \textit{JITTER}, is the victim in the peer bullying scenario; and \textit{FUSE}, is the bully (See Fig. \ref{fig:REMindCharacters}). During the game, players observe three Furhat robots enact a bullying conflict, use a physical ``Bully-Detector'' prop (see \cite{sanoubari2024makes}) to properly identify the aggressive behavior, and complete an empathy training task to understand the emotional perspectives of the characters. The experience culminates in ``Puppet Mode,'' where the child records an improvised intervention and puppeteers \textit{AVATAR} to change the outcome of the live robotic drama, allowing them to safely witness the outcomes of bystander intervention.

The game was designed through a highly iterative co-design process that involved diverse stakeholders (including children, teachers, and domain experts) in various phases of development over the span of 4 years. The final stage of prototyping involved conducting technical rehearsals with a theater director over 3 months~\cite{sanoubari2026aesthetics}, which resulted in a 2-hour interactive experience. To handle the emotional nuance and technological complexity of the game, the system is developed semi-autonomously, and experience is seamlessly orchestrated by a hidden technical operator (the Wizard) and an in-room drama facilitator (the Joker). Because \textit{REMind} is a complex system, this section describes only the system elements that are directly relevant to the learning goals and evaluation reported in this paper.

\subsection{Technical Implementation}
\textit{REMind} is set up as a room-based installation (see Fig. \ref{fig:REMindScene}) using \textit{Furhat} robots: a tabletop social robot platform with speech, head movement, and a back-projected animated face (capable of displaying animated life-like facial expressions)~\cite{al2012furhat}. To script the experience, we developed StorySync, a custom spreadsheet-based toolkit designed to synchronize digital system components (including multiple Furhat robots, a graphical user interface (GUI), and ambient audio-visual cues). StorySync reads the script from a spreadsheet line-by-line and automatically generates dynamic buttons on a control screen, allowing the Wizard to trigger pre-scripted robot actions, dialogue, and environmental changes.

To allow players to ``puppeteer'' \textit{AVATAR}, the system captures their improvised performance using  `Live Link Face' iOS application (developed by Epic Games~\cite{epicgames_livelinkface}), which leverages Apple's ARKit to record high-fidelity facial motion data. This data is exported and processed through Furhat's Gesture Capture tool~\cite{furhat2025gesturecapture}, which converts the human facial parameters into gesture files that the robot can execute. Meanwhile, the player's voice is extracted directly from the same video recording and converted into an audio file.

\subsection{Learning Goals}
\label{learning-goals}
\textit{REMind} pursues specific learning goals that were determined based on interviews with local educators~\cite{sanoubari2022designing}, and consultations with social psychologists (domain experts in peer bullying). Among these goals, are the following:
{
\renewcommand{\labelenumi}{\textbf{[G\arabic{enumi}]}}
\begin{enumerate}
    \item \textbf{Cultivate Empathy and Compassion} by reflecting on victims’ feelings and perspectives.
    \item \textbf{Recognize Bully's Underlying Motivations}, such as insecurity, or the desire for power and control.
    \item \textbf{Practice Defending Strategies}, specifically, by confronting the bully, or comforting the victim.
    \item \textbf{Understand Outcomes of Defending}, including the impact on the bully, who may or may not stop bullying, and the victim, whose suffering will be alleviated.
    \item \textbf{Reflect on Self-Efficacy} and build confidence in their own ability to intervene and support peers.
\end{enumerate}

It is worth noting that \textit{REMind} also had other learning objectives that fall outside of the scope of this manuscript.

\subsection{Interactive Narrative}
\label{5_phases}
\textit{REMind} unfolds as a five-phase interactive narrative in which the child helps a robotic avatar respond to a bullying incident. Below is a summary of the parts of the experience most relevant to the findings reported in this paper.

\textbf{Phase 1: Entering the Game.} The child customizes \textit{AVATAR} by choosing its name, appearance, and accessories (a hat, and a small plush toy). When child initiates the game, the room lights transition, and \textit{AVATAR} wakes up, introduce itself, and explain that a mysterious ``glitch'' is causing robots to malfunction (such as acting unkind or forgetting their friends), and asks for help.

\textbf{Phase 2: Simulating Peer Bullying.}
To investigate the glitch, the child must ``simulate'' one of \textit{AVATAR}'s past memories, which causes the other two robots to activate and enact a scene where \textit{FUSE} bullies \textit{JITTER} by using verbal mockery and social exclusion, and \textit{AVATAR} watches silently. The system then indicates that \textit{AVATAR}'s ``Bully-Detector'' is broken, tasking the child with exploring and fixing the issue, teaching the robot to recognize the situation as bullying.

\textbf{Phase 3: Empathy and Perspective-Taking.}
Next, the child helps \textit{AVATAR} fill in the gaps of its understanding, by assisting the robot with developing empathy and taking the perspective of other characters. First, they 
reflect on \textit{JITTER}’s emotional state by interpreting an affective display using light and music cues~\footnote{Since data and analysis from this module was substantial in its own right, it has been excluded from the present manuscript.}, and then, they reflect on the thoughts of the victim, and motivations of the bully. 

\textbf{Phase 4: Puppet Mode --- Rehearsing Intervention.}
The game then presents two defending strategies for \textit{AVATAR}: \textit{Comfort JITTER} or \textit{Be Firm with FUSE}. After selecting one, the child improvises how \textit{AVATAR} should enact this strategy, and their performance is recorded. Then, they use the \textit{Puppet Mode} to re-simulate the scene, and replay the recorded intervention on the robot, allowing \textit{AVATAR} to react to the bullying scene as a bystander, using the child's own recorded facial expressions and voice.

\textbf{Phase 5: Outcomes and Resolution.}
In the final phase, the child explores the consequences of intervention. As the narrative unfolds, the child sees that confronting the bully can unfold in more than one way: in one version, \textit{FUSE} backs down, while in another, \textit{FUSE} retaliates. The game then shifts to \textit{JITTER}’s perspective, contrasting the victim’s distress before intervention with their relief afterward, showing that even when bullying is not fully stopped, peer support can still make the victim feel better. The story then concludes with the resolution of the glitch.

\section{METHODOLOGY: PLAY-TESTING STUDY}
The play-testing study set out to evaluate whether \textit{REMind} supported the intended learning objectives. This section provides an overview of the experiment design, with a focus on the goals outlined in Section \ref{learning-goals}.

\subsection{Setup}
The installation took place in a university room staged as a small studio apartment (Fig. \ref{KERNELSetup}). The space was organized into three functional areas: (1) an area for onboarding and avatar customization, (2) a main gameplay area containing the robots, and (3) a hidden wizarding station separated by room dividers. The Wizard was able to see the interactions in real-time via a 360-degree camera. The three robots were arranged in a triangle on a table in the main gameplay area. 

\subsection{Measurements}
We collected pre/post quantitative data from self-report scales that assessed self-efficacy for defending, empathy, and outcome expectations using adapted items from P{\"o}yh{\"o}nen et al.~\cite{poyhonen2013defending}. More specifically, self-efficacy was measured with an item asking \textit{``How easy or hard would it be for you to defend the victim”}. Outcome expectations regarding the bully and the victim were measured with two items asking \textit{``Do you think the bully would listen if you told them to stop,''} and \textit{``Do you think telling the bully to stop would help the victim feel better''}), respectively.
Empathy scores towards the victim  were measured using parallel items referring to a human victim at pre-test and the robot at post-test (i.e., \textit{``When the bullied person/robot was sad, I also felt sad”}). 

Second, we collected structured game logs including participants' selected bystander intervention strategy, as well as open-ended reflection to prompts that were used to capture their empathic reaction to the bullying scene, and elicit perspective-taking of different characters. Immediately after watching the multi-robot dramatization of peer bullying, these prompts asked: \textit{``How do you feel,''} \textit{``Why do you think your AVATAR did not do anything''}. In phase 3, the prompts asked, \textit{``If we could hear JITTER’s thoughts, what would we hear,''} and, \textit{``Why do you think FUSE bullied JITTER?''}

Third, we collected semi-structured interview data on children's knowledge of defending strategies before and after gameplay, as well as reflections shared during the collective debriefing.
The Joker recorded interview responses, gameplay observations, and debriefing notes, while children responded to self-report items directly using a sliding-scale.

\subsection{Procedure}
For recruitment, we installed posters in public spaces, with a QR code that parents could scan to receive information about the study and sign up by providing parental consent, and obtaining assent from their child. Participants were informed that they will play a make-believe anti-bullying game with social robots to help the researchers
evaluate it, and that the game involves pretending to help a programmer (a researcher) fix a robot that’s having a problem.

\subsubsection{Single-player Play-testing}
Each child arrived at the study site at a scheduled time, was greeted by a researcher (the Joker), and brought into the experiment room (see Fig. \ref{KERNELSetup}).
After confirming child's assent, the Joker administered a brief pre-test semi-structured interview, in which participants were asked to recall the last time they had witnessed bullying, describe how they had responded, and answer follow-up questions probing their knowledge of defending strategies. This was followed by administering self-report items on empathy, self-efficacy for defending, and outcome expectations.

\begin{figure}[!ht]
  \centering
  \framebox{%
    \includegraphics[
      width=0.95\linewidth,
      clip
    ]{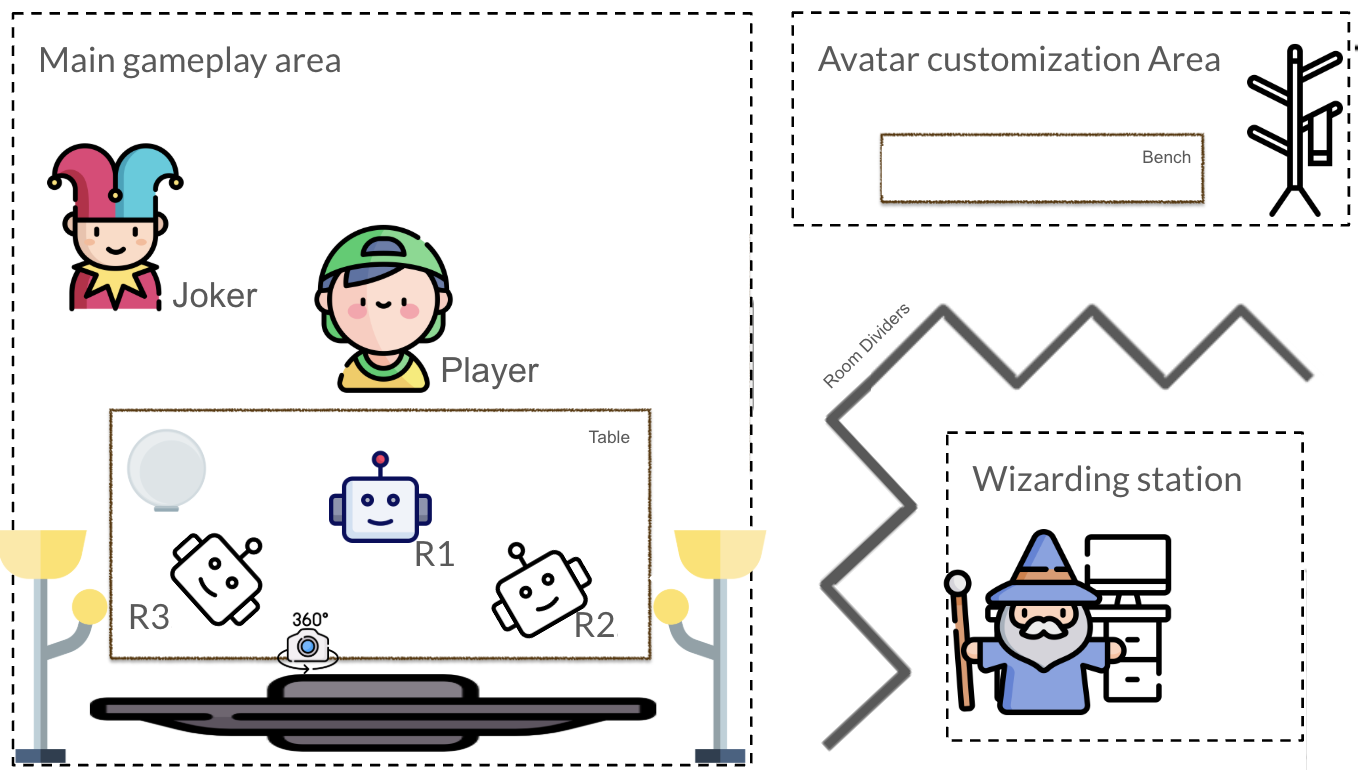}%
  }

  \captionsetup{aboveskip=2pt, belowskip=2pt}
  \caption{Experiment room layout.}
  \label{KERNELSetup}
\end{figure}

The child then moved to the main gameplay area to play \textit{REMind} as outlined in Section \ref{5_phases}. The game was paused at designated moments to collect in-situ qualitative data. After the bullying memory in Phase 2, the Joker captured participant emotional reaction to the scene and their interpretation of underlying reasons for \textit{AVATAR}'s inaction as bystander. During Phase 3, the Joker recorded participant perspective-taking reflections about \textit{JITTER} and \textit{FUSE}. In Phase 4, participant choice of  defending strategy for \textit{AVATAR} was logged as a game log. 
At the end of Phase 5, \textit{AVATAR} gifted the plush toy (from Phase 1) to the child, as a token of appreciation, before saying goodbye.

After gameplay, the Joker administered the post-test measures including a semi-structured interview and mirrored self-report items. In the post-test interview, participants were asked them to imagine a future peer bullying scenario and reflect on how they might respond, to re-assess their knowledge of defending strategies. The post-test self-report items mirrored the same constructs as the pre-test.

\subsubsection{Collective Debriefing}
Parents were invited to enroll their child in a later group debriefing session, scheduled up to a month after the initial play-testing.
In this session, children reflected on the experience together, discussed what they remembered from the game, and shared whether they had encountered bullying in the weeks following the play-test and how they responded. The session ended with a behind-the-scenes demonstration of how \textit{REMind} was orchestrated the Wizard of Oz (WoZ) technique.

This protocol was approved by University of Waterloo Research Ethics Committee.

\subsection{Participants}
We recruited 18 children aged 9--10 years (M= 9.22, SD = 0.43). This target age range was recommended by expert psychologists based on their prior experiences with the KiVa Anti-Bullying program~\cite{herkama2018kiva}. In verbal and written communication, parents referred to 11 children using he/him pronouns and 7 using she/her pronouns. Among those, 12 children participated in the group debriefing sessions. The remaining 6, who were unable to attend due to scheduling constraints, received debriefing letters written in simple language instead.

\subsection{Analysis}
The quantitative analysis focused on three pre-registered expectations derived from \textit{REMind}’s learning goals:
{
\renewcommand{\labelenumi}{\textbf{[H\arabic{enumi}]}}
\begin{enumerate}
    \item \textbf{Self-efficacy:} children’s perceived self-efficacy for defending would increase following the intervention.
    
    \item \textbf{Outcome expectations:} (a) expectations that defending would stop the bullying would shift toward a calibrated midpoint, reflecting the partial success modeled in the story, and (b) expectations that defending would help the victim feel better would increase.
    
    \item \textbf{Empathy:} (a) children’s empathy towards the victim would increase following the intervention, (b) empathy toward the robot avatar would be positively associated with baseline empathy toward a human victim.
\end{enumerate}

Self-efficacy gain (H1) was tested using paired-samples t-tests on pre/post measurements. For outcome expectations, we defined normative target values for each item: 50 (out of 100) for bully-related expectation (reflecting that bullying may or may not stop as a result of defending) and 100 for victim-related expectation (reflecting that defending has a consistently positive effect in how the victim feels). We calculated each participant’s absolute distance from the target value at pre- and post-test. A one-tailed paired t-test was used to assess whether this distance significantly decreased following the intervention, indicating movement toward the expected interpretation of the narrative outcomes (H2). Finally, we used a Spearman correlation to test the association between human-directed and robot-directed empathy (H3).

Qualitative data were analyzed using inductive thematic analysis. \textbf{Intercoder Reliability} was established by a second coder on 25\% of the data (a threshold recommended by the literature~\cite{o2020intercoder}), yielding 94\% agreement and Cohen’s $\kappa = 0.93$, which represents a near-perfect agreement.

\section{RESULTS}

\subsection{Quantitative Findings} 

\subsubsection{Self-Efficacy}
Children's perceived self-efficacy significantly increased from pre-test (M = 39.2, SD = 20.5) to post-test (M = 53.9, SD = 21.3), t(17) = 3.12, p = .003, following the intervention. Cohen’s d = 0.74 indicated a medium-to-large effect size. See Fig. \ref{fig:self-efficacy}.

\begin{figure}[htbp]
  \centering
  \framebox{%
    \includegraphics[width=0.95\linewidth]{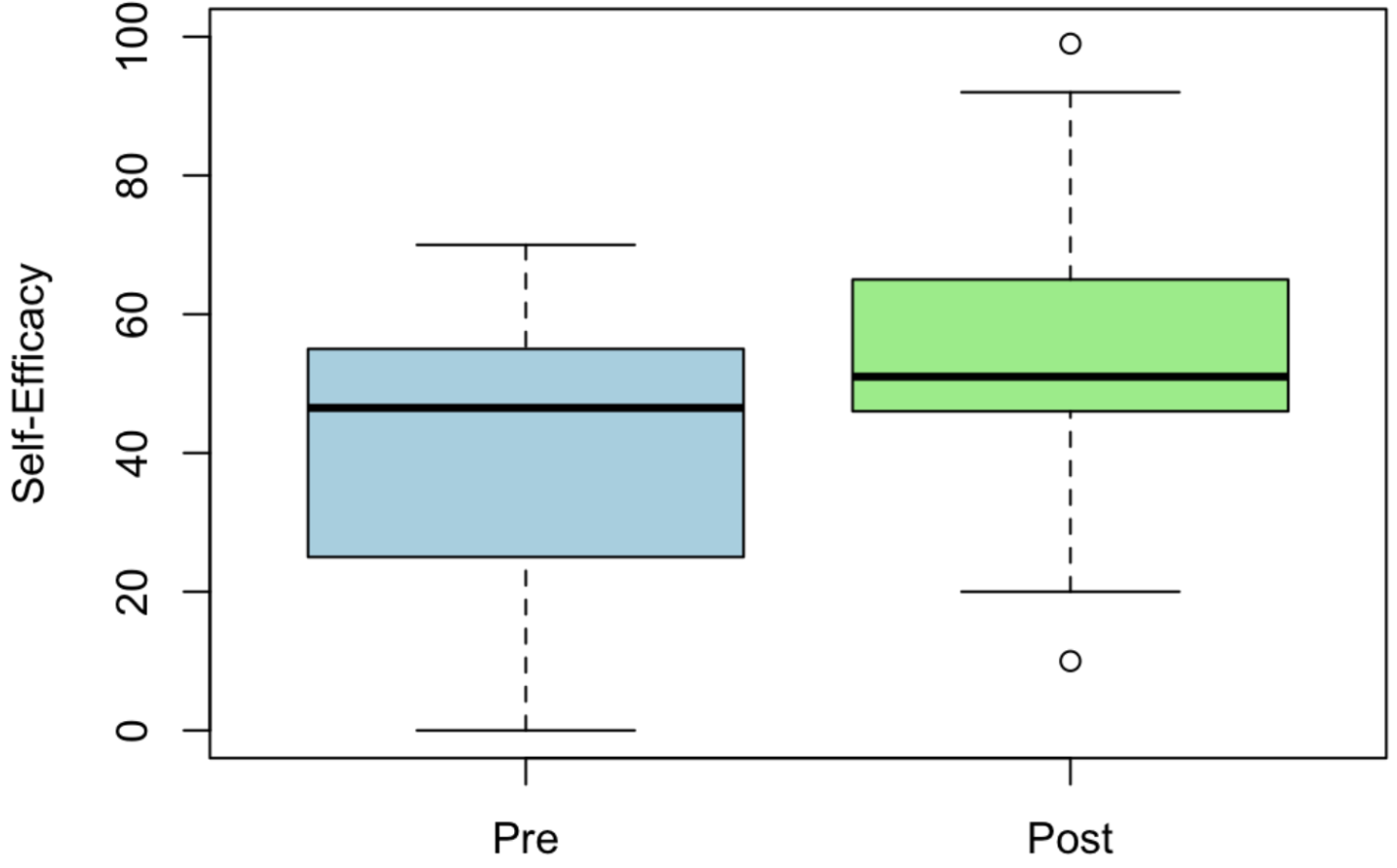}%
  }
  \caption{Self-efficacy for defending increased post-intervention. Boxes show the interquartile range (IQR); whiskers extend to 1.5× IQR.}
  \label{fig:self-efficacy}
\end{figure}

\subsubsection{Outcome Expectations}
For the bully-related item (target = 50), expectation scores decreased from M = 52.22 to 47.44. A one-tailed paired t-test on distance from target showed a significant reduction, t(17) = –3.11, p = .003.  
This confirmed H2(a).
For outcome expectations regarding the victim (target = 100), scores increased from M = 70.94 to 78.06, but the change was not significant: t(17) = –0.89, p = .193. Therefore, H2(b) was not confirmed. See Fig. \ref{fig:outcome-expectations}.

\begin{figure}[h!]
  \centering
  \framebox{%
    \includegraphics[width=0.95\linewidth]{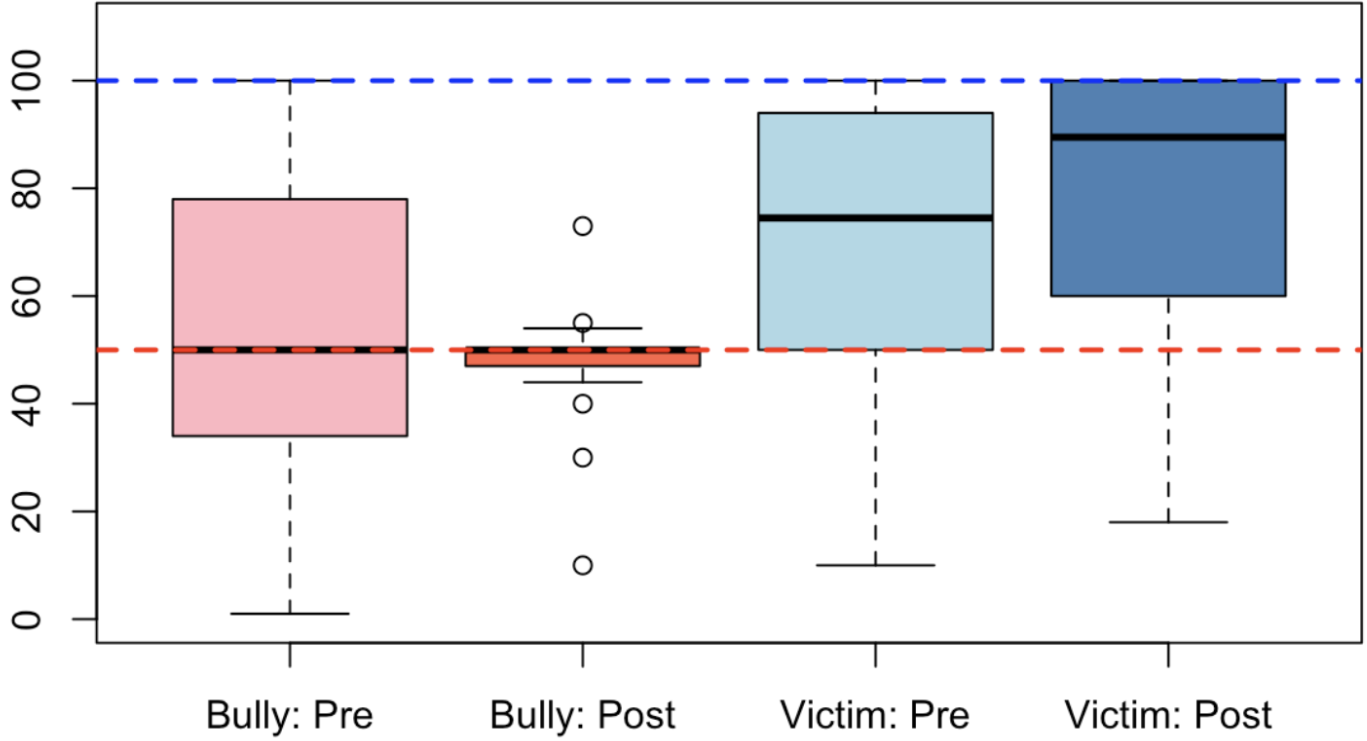}%
  }
  \caption{Post-test responses reflected more accurate expectations about defending outcomes. Red and blue lines mark bully- and victim-related targets, respectively.}
  \label{fig:outcome-expectations}
\end{figure}

\subsubsection{Empathy}
Empathy scores increased by 15.7 points, but the change was not statistically significant, t(17) = 1.62, p = .062. H3(a) was not confirmed.
Furthermore, a significant positive correlation was found between empathy toward the human character at pre-test and the robot avatar at post-test, Spearman’s $\rho$ = .62, p = .006, confirming H3(b), this indicates that children with higher baseline empathy were more likely to empathize with the robot. See Fig. \ref{fig:empathy-combined}.

\begin{figure}[htbp]
  \centering
  \begin{subfigure}[b]{\linewidth}
    \centering
    \framebox{%
      \includegraphics[width=0.95\linewidth]{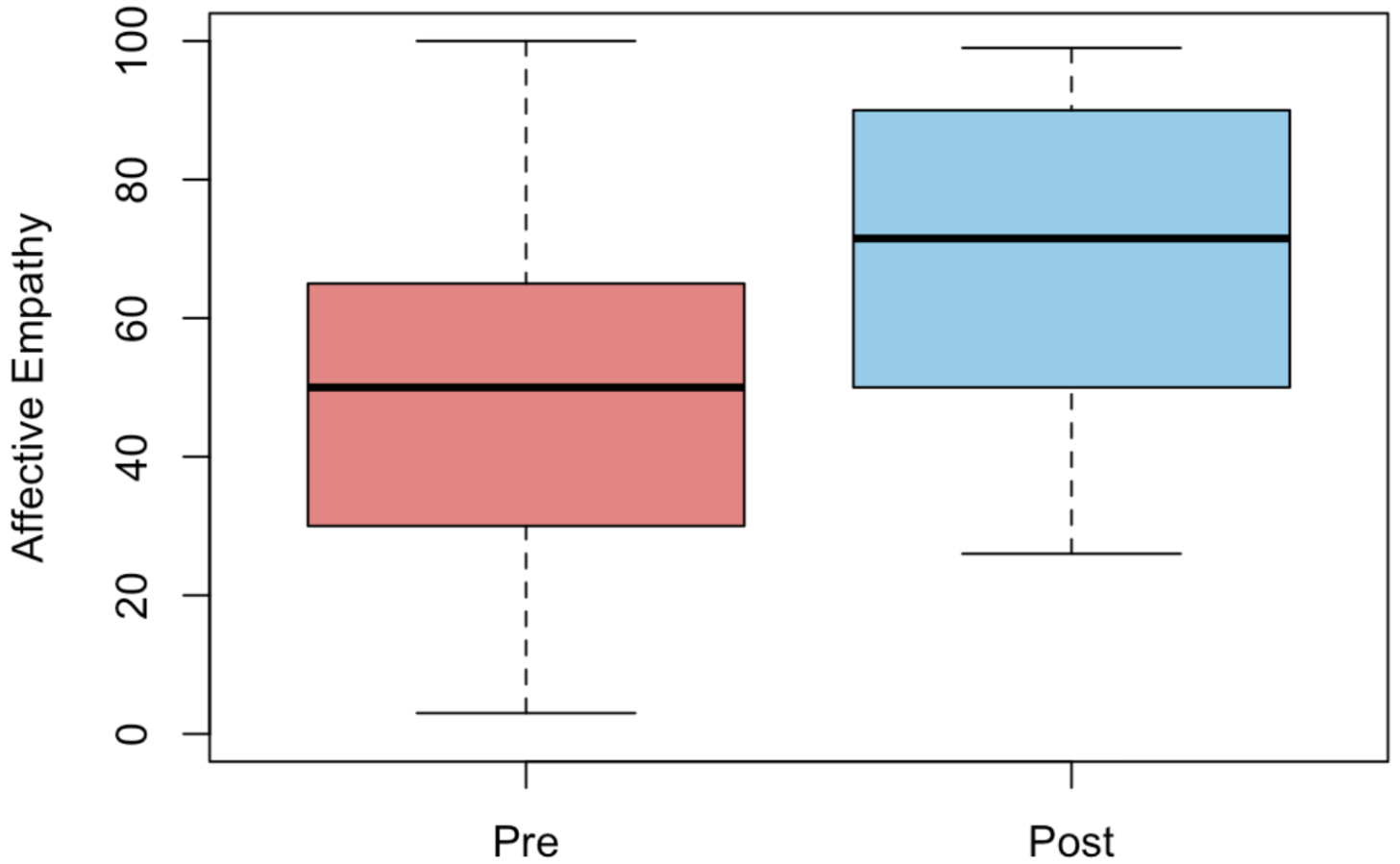}%
    }
    \caption{}
    \label{fig:affective-empathy}
  \end{subfigure}

  \vspace{2pt}

  \begin{subfigure}[b]{\linewidth}
    \centering
    \framebox{%
      \includegraphics[width=0.95\linewidth]{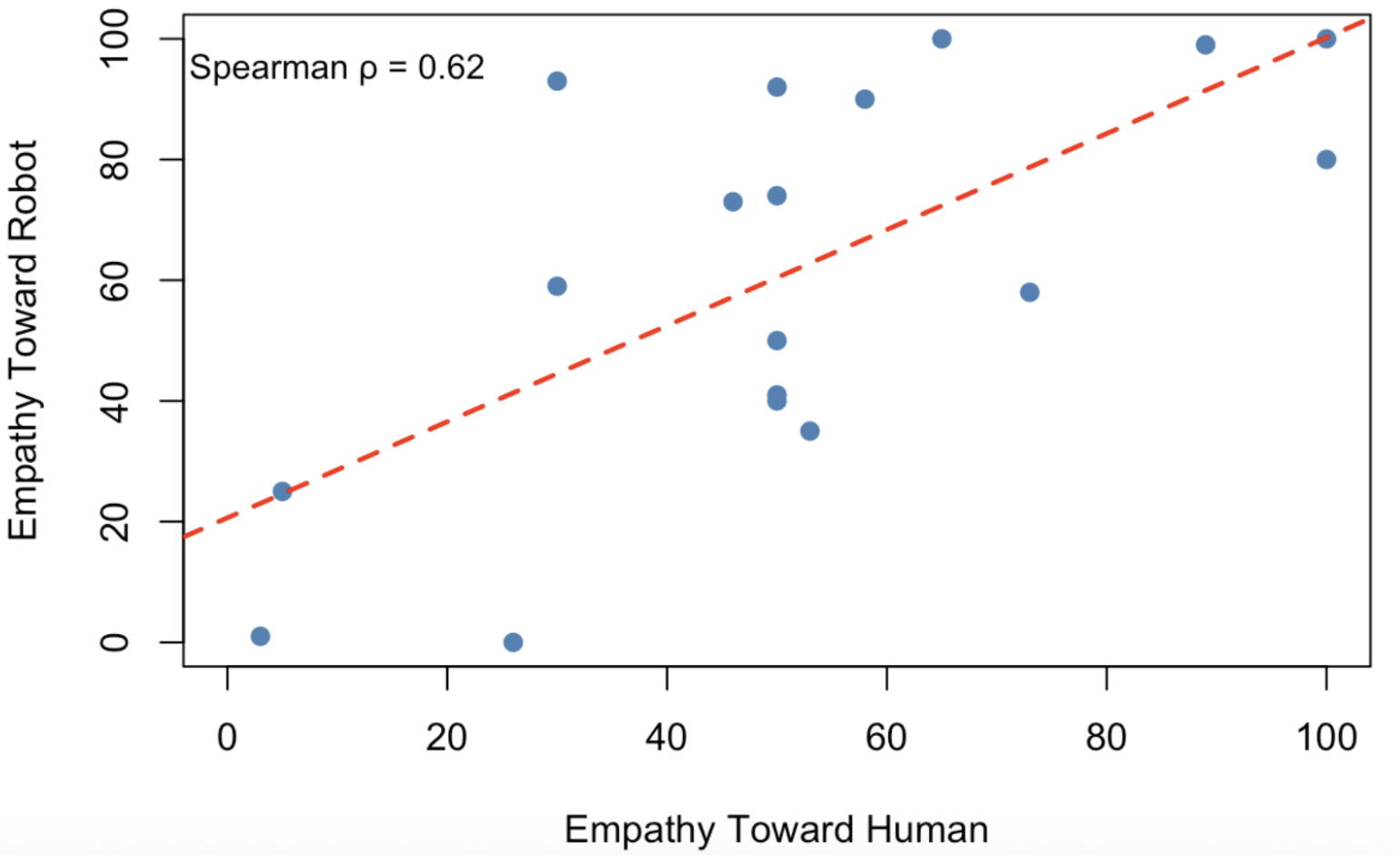}%
    }
    \caption{}
  \end{subfigure}

  \caption{(a) Empathy scores before and after the intervention. No significant change was observed. (b) Relationship between empathy toward humans and the robot avatar.}
  \label{fig:empathy-combined}
\end{figure}

\subsection{Qualitative Themes: Empathy \& Perspective-Taking}
\label{sec:qual_results}
Inductive coding of children's responses to `How do you feel' (asked right after watching the bullying scene) revealed that 94\% of participants (17 out of 18) felt sad for \textit{JITTER} (victim), while one child responded with `I don't know'.  
Also, 16\% were both sad for \textit{JITTER} (victim) and mad at \textit{FUSE} (bully). 
For example, one participant replied: ``That guy shouldn't be bullying;'' then, pointing to \textit{JITTER} (victim) they said ``I feel bad for him,'' pointing to \textit{FUSE} (bully) they added ``and mad at him'' (see Fig. \ref{fig:how-do-you-feel}).

\begin{figure*}[ht!]
    \centering
    \framebox{%
      \includegraphics[width=0.95\linewidth]{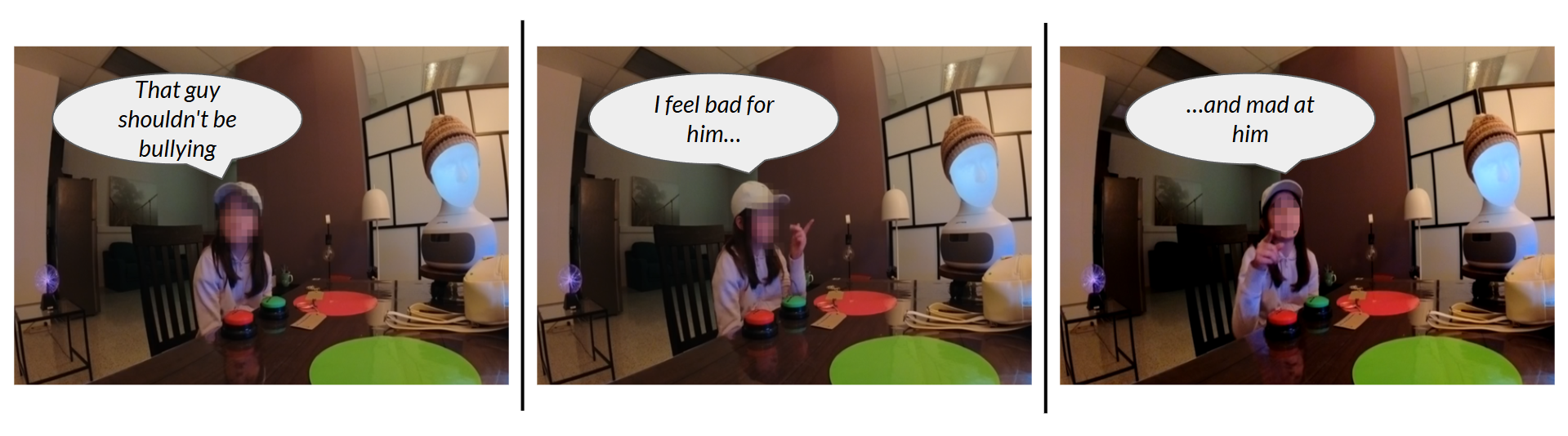}%
    }
    \captionsetup{aboveskip=2pt, belowskip=2pt}
    \caption{Player interaction in \textit{REMind} demonstrating an example of multi-robot drama facilitating perspective-taking.}
    \label{fig:how-do-you-feel}
\end{figure*}

\subsubsection{Bystander's Perspective}
When reflecting on potential reasons for \textit{AVATAR}'s inaction as a bystander, over half (55\%) attributed it to low skills or confidence (e.g.,\textit{ ``maybe she didn't feel confident, because of FUSE's rude words.''} [P9], or \textit{``she doesn't know how to stand up.''} [P2]). Responses from 38\% of children indicated fear of bully's retaliation (e.g., \textit{``hmmmm[...] because he also didn't wanna get bullied'' [sic]}[P18]. 16\% hinted at low peer status, e.g., \textit{``because he wasn't popular and he didn't think he can get out of this situation. ''} [P8]). Finally, 11\% indicated that avatar was indifferent (e.g., \textit{``because he didn't really get involved with the problem''} [P5]).

\subsubsection{Victim's Perspective}
Two overarching themes emerged from analyzing children’s improvised responses on what the Victim was thinking.
The first theme reflected the victim responding to and coping with the bullying.
Half of the participants indicated that the victim wishes for the bullying behavior to end, imagines fighting back, talking back, or expressing anger toward bully (e.g., \textit{``stop, I don't like what you're doing, just stop''} [P12], \textit{``why are you bullying me?''} [P10]).
27\% of players indicated that the victim has a desire to escape the situation, wants to leave, disappear, or avoid the bullying context (e.g., \textit{``get me out of this place!''} [P4], \textit{``oh I wish I wasn't here today, oh my god I want to disappear''} [P13], \textit{``I never want to go back to school''} [P14]).
Finally, 16\% of participants imagined the victim hopes or wishes that a bystander would intervene, support, or help (e.g., \textit{``Bob! please help me! or someone! at least somehow!''} [P6], \textit{``I wish someone can stand up for me''} [P17], \textit{``William [avatar robot], why wouldn't you stand up for me''} [P14]).

The second emerging theme focused on the victim's internalized emotional experience. Responses from 38\% of participants suggested victim has negative self-image, low self-worth, shyness, self-blame, or feeling unwanted or alone (e.g., \textit{``I must be so shy [...] I must be so weird [...] nobody likes me'' }[P2], \textit{``I'm not important[...] I'm alone''} [P1], \textit{``I want to talk louder, but I can't, I'm too shy'' }[P11], \textit{``I have no friends''} [P16]).
Furthermore, one third of participants imagined the victim expressing emotional distress, feeling sad, scared, overwhelmed, or upset (e.g., \textit{``I'm sad and also scared'' }[P9], \textit{``I don’t feel safe'' }[P12]).

\subsubsection{Bully's Motivation}
When asked to reflect on why \textit{FUSE} may have acted the way he did, children offered a range of interpretations. 
Some demonstrated awareness of the power and social dynamics at play. 38\% of participants described \textit{FUSE} as motivated by seeking social dominance, power, or attention from others (e.g., \textit{``because FUSE wanted lots of attention and wanted JITTER to feel ashamed''} [P2], \textit{``cuz he wanted to make the crowd laugh, and to make JITTER feel bad''} \textit{[P11]}, \textit{``I don’t think this is a good reason, but I think he wanted to be popular''} [P15]). 27\% of responses emphasized that \textit{JITTER} was an easy target because of perceived vulnerabilities (e.g., \textit{``because he knew JITTER didn’t have any friends''} [P3], \textit{``either he wasn’t having a good day, or he just saw JITTER was very shy and he felt like JITTER was an easy target''} [P13]).

The second emerging theme reflected bully's moral disengagement. 27\% of responses indicated that \textit{FUSE} bullies due to personal struggles, or family problems (e.g., \textit{``because FUSE was sad. something was going on in their life''} [P6], \textit{``maybe he is having rough moments, and his father is sick, he’s taking his anger on JITTER''} [P9], \textit{``Most adults will say, oh , his parents aren't very nice, or something at home isn't very good for him''} [P10]. 
16\% suggested that \textit{FUSE}’s behavior was linked to \textit{JITTER}'s perceived flaws (e.g., \textit{``because FUSE thinks that JITTER is bad at everything''} [P16], \textit{``FUSE bullied JITTER because he looked really plain''} [P18]).
11\% framed it as entertainment or joking (e.g., \textit{``FUSE wanted to joke; but they weren’t kind and [JITTER] didn’t like it, but FUSE thought it was just a joke''} [P7]). 
Finally, a few responses (2 out of 18) offered essentialist explanations, portraying \textit{FUSE}’s behavior as an inherent trait rather than situational (e.g., \textit{``because FUSE is a bully''} [P7]), and one player expressed uncertainty.

\subsection{Descriptive Statistics: Bystander Intervention}
Among the two choices of defending strategies, only 4 participants selected the `Comfort the Victim' strategy.
On average, participants who chose to comfort the victim reported higher perceived self-efficacy for their chosen action (M = 64.5, SD = 25.2) compared to those who chose to confront the bully (M = 42.6, SD = 21.1). They also reported higher expectations that their action would help the victim (M = 95.0, SD = 10.0 vs. M = 74.2, SD = 27.1). Participants who chose to confront the bully reported higher expectations that the bully would stop (M = 59.5, SD = 20.4 vs. M = 40.0, SD = 22.0). These descriptive evidence should be interpreted with caution due to the severe group size imbalance.

Also, 14 out of 18 participants named at least one new intervention strategy post-test (such as one of the choices introduced in the game), with one participant specifically stating \textit{``I would say exactly what I told $<$\textit{RobotName}$>$ to say''} [P7]. Finally, among the twelve children who participated in the collective debriefing session, at least four told the group that they stood up to a bully in real-life after playing the game, two of whom emphasized that they used the same words they had improvised during the gameplay.

\section{DISCUSSION}
\subsection{Revisiting Design Goals: Summary of Learning Gains}
\textit{REMind} showed evidence of supporting all five learning goals outlined in Section \ref{learning-goals}, although the strength of evidence varied across goals and measures.
For \textbf{empathy and compassion} [G1], although the quantitative increase in empathy was not statistically significant, the qualitative findings suggest meaningful empathic engagement during play. Nearly all participants reported feeling sad for \textit{JITTER}, and many articulated the victim’s fear, loneliness, and desire for help. Regarding \textbf{recognizing the bully’s underlying motivations} [G2], children frequently reflected on \textit{FUSE}’s behaviour in terms of power, attention, social dynamics, or personal struggles, than simply “mean” behavior. 
For \textbf{practicing defending strategies} [G3], all participants rehearsed bystander intervention using one of the two intervention strategies offered in the game through Puppet Mode. In terms of \textbf{understanding outcomes of defending} [G4], children’s expectations became significantly better calibrated on the bully-related item, aligning more closely with the idea that confronting a bully may or may not stop the bullying. Expectations that defending would help the victim remained high, consistent with the intended message that intervention can still provide meaningful support even when it does not fully stop the aggressor. And finally, participants' perceived \textbf{self-efficacy} [G5] increased significantly from pre- to post-test, and many children explained the robotic avatar’s inaction in terms of low confidence or fear of retaliation, indicating that it had become a salient object of reflection during play. Overall, \textit{REMind} appears to have been not only engaging, but pedagogically meaningful as an RMAD experience.

\subsection{Trade-Offs and Limitations}
While our findings offer insights into the design and evaluation of robot-mediated, drama-based anti-bullying interventions, the study is subject to several limitations that constrain the generalizability and interpretability of the results. First, the study design privileged experiential richness and participant responsiveness over strict experimental control. Aspects of the intervention, were intentionally kept flexible to accommodate participant engagement, self-reflection and pedagogical flow—for example, the Joker used improvisation to address unscripted questions by children. While this is consistent with the ethos of drama education, it inevitably introduces variability that limits reproducibility. Also, the complex nature of this system, makes it difficult to isolate the causal effects of specific design elements. As such, outcomes observed, cannot be straightforwardly attributed to any single design component. In other words, as designers of REMind, we prioritized children's experiential learning and pedagogical richness over strict experimental control.

Furthermore, embedding a researcher as the in-room facilitator (the Joker) introduced the risk of bias. To mitigate this, we asked participants to respond to self-report measures directly using sliding scale questions, and conducted intercoder reliability testing to ensure their verbalizations and actions were not qualitatively interpreted in a biased manner.

Finally, the study did not compare the \textit{REMind} system to traditional, non-robotic anti-bullying interventions, making it impossible to claim that the robotic system is uniquely superior. Also, the study only captured immediate reactions and short-term reflections during the two-hour session and subsequent debriefing sessions, leaving it unclear whether the intervention causes durable behavioral changes in real-world bystander situations. We acknowledge that addressing this gap requires future longitudinal research.

\subsection{Towards Long-term Real-World Deployment}
REMind's scalability is constrained by its semi-autonomous nature, and the two-person WoZ facilitation protocol.
To enable real-world adoption, future work must explore and mitigate the practical barriers for long-term deployment of such systems in educational setting. 
In facilitating \textit{REMind}, we argue that the role of the joker should ideally not be automated. \textit{REMind} is modeled after Forum Theatre~\cite{boal2000theater}, which Joker plays a key facilitator role by guiding reflection and encouraging participation ~\cite{sanoubari2026aesthetics}. In real-world use, a drama educator may take the joker role; as \textit{REMind} is meant to augment, not replace, educators. The wizard however, controls the narrative flow and their role should ideally be fully automated. Future work could explore simplified teacher-facing interfaces (e.g., clicker-based story progression) and the use of pre-trained large language models (LLMs) to support light narrative branching and dialogue handling. Also, future work should work with educators to identify other barriers, functional needs, and opportunities for deployment in educational settings.

\section{CONCLUSIONS}
This paper presented \textit{REMind}, an educational robot-mediated role-play game designed to help children rehearse bystander intervention in bullying situations. Through a mixed-methods play-testing study with 18 children, we found evidence that the experience supported several of its intended pedagogical goals. Quantitatively, children showed a significant increase in perceived self-efficacy for defending, and their expectations about whether confronting a bully would stop the bullying became better calibrated to the game's modeled outcomes. Qualitatively, children demonstrated strong empathic engagement with the victim, reflected on the bystander's fear and uncertainty, and articulated nuanced interpretations of the bully's motivations.
Taken together, these findings suggest that \textit{REMind} offers more than a novel interface for anti-bullying education. By combining embodiment, narrative, and rehearsal, it creates a space in which children do not merely receive advice about defending peers, but actively practice what defending might feel like, what it might accomplish, and why it can still matter even when outcomes are uncertain. More broadly, this work contributes evidence that Robot-Mediated Applied Drama (RMAD) can support social-emotional learning by positioning robots not only as tutors, but as embodied dramatic proxies for reflection and action.
At the same time, this study reports an early-stage evaluation of a complex, semi-autonomous research prototype. Future work should examine longer-term effects, and explore how such systems can be adapted for sustained use in real educational settings. Even with these limitations, the present findings indicate that robot-mediated role-play is a promising direction for designing interactive systems that support prosocial action in socially difficult situations.

\section*{ACKNOWLEDGMENT}
We are grateful to psychologists Dr.\ Christina Salmivalli and Dr.\ Claire Garandeau (University of Turku) for expertise that informed the learning goals, target audience, and measurements of \textit{REMind}. We also thank the Centre for Spectatorship and Audience Research (University of Toronto) for the feedback that shaped the experiment design, particularly on the importance of including the Joker as a Forum Theatre facilitator and the value of collective debriefing sessions.

\bibliographystyle{IEEEtran}
\bibliography{references}

@misc{sanoubari2026aesthetics,
    title = {{Aesthetics of Robot-Mediated Applied Drama: A Case Study on REMind}},
    year = {2026},
    author = {Sanoubari, Elaheh and Pan, Alicia and Rebello, Keith and Fernandes, Neil and Houston, Andrew and Dautenhahn, Kerstin},
    url = {https://arxiv.org/abs/2603.23816},
    arxivId = {cs.HC/2603.23816}
}

@article{salmivalli2005anti,
    title = {{Anti-bullying intervention: Implementation and outcome}},
    year = {2005},
    journal = {British journal of educational psychology},
    author = {Salmivalli, Christina and Kaukiainen, Ari and Voeten, Marinus},
    number = {3},
    pages = {465--487},
    volume = {75},
    publisher = {Wiley Online Library}
}

@article{poyhonen2013defending,
    title = {{Defending behavior in bullying situations}},
    year = {2013},
    author = {P{\"{o}}yh{\"{o}}nen, Virpi and {others}},
    publisher = {Annales Universitatis Turkuensis B 367}
}

@inproceedings{sanoubari2022designing,
    title = {{Designing an Anti-Bullying Serious Game: Insights from Interviews with Teachers}},
    year = {2022},
    booktitle = {Joint Intl. Conf. on Serious Games (JCSG)},
    author = {Sanoubari, Elaheh and Cardona, John E Muñoz and Houston, Andrew and Young, James and Dautenhahn, Kerstin},
    pages = {102--121},
    organization = {Springer}
}

@article{bowman2014educational,
    title = {{Educational live action role-playing games: A secondary literature review}},
    year = {2014},
    journal = {The Wyrd Con Companion Book},
    author = {Bowman, Sarah Lynne},
    pages = {112--131},
    volume = {3},
    publisher = {Wyrd Con Los Angeles, CA}
}

@article{garandeau2023effects,
    title = {{Effects of the KiVa anti-bullying program on defending behavior: Investigating individual-level mechanisms of change}},
    year = {2023},
    journal = {Journal of school psychology},
    author = {Garandeau, Claire F and Turunen, Tiina and Saarento-Zaprudin, Silja and Salmivalli, Christina},
    pages = {101226},
    volume = {99},
    publisher = {Elsevier}
}

@misc{epicgames_livelinkface,
    title = {{Epic Games: Live Link Face}},
    year = {2024},
    url = {dev.epicgames.com/documentation/en-us/unreal-engine/live-link-face-device}
}

@article{prager2019exploring,
    title = {{Exploring the use of role-playing games in education}},
    year = {2019},
    journal = {The MT Review},
    author = {Prager, Richard Heinz Patrick}
}

@inproceedings{hall2009fearnot,
    title = {{FearNot!: providing children with strategies to cope with bullying}},
    year = {2009},
    booktitle = {Proceedings of the 8th International Conference on Interaction Design and Children},
    author = {Hall, Lynne and Jones, Susan and Paiva, Ana and Aylett, Ruth},
    pages = {276--277},
    organization = {ACM}
}

@misc{furhat2025gesturecapture,
    title = {{Furhat Robotics: Gesture Capture}},
    year = {2025},
    url = {docs.furhat.io/gesture_capture_tool/}
}

@inproceedings{al2012furhat,
    title = {{Furhat: a back-projected human-like robot head for multiparty human-machine interaction}},
    year = {2012},
    booktitle = {Cognitive Behavioural Systems},
    author = {Al Moubayed, Samer and Beskow, Jonas and Skantze, Gabriel and Granstr{\"{o}}m, Björn},
    pages = {114--130},
    organization = {Springer}
}

@article{o2020intercoder,
    title = {{Intercoder reliability in qualitative research: debates and practical guidelines}},
    year = {2020},
    journal = {International journal of qualitative methods},
    author = {O’Connor, Cliodhna and Joffe, Helene},
    volume = {19},
    publisher = {SAGE Publications Sage CA: Los Angeles, CA}
}

@incollection{herkama2018kiva,
    title = {{KiVa antibullying program}},
    year = {2018},
    booktitle = {Reducing cyberbullying in schools},
    author = {Herkama, Sanna and Salmivalli, Christina},
    pages = {125--134},
    publisher = {Elsevier}
}

@incollection{hammer2018learning,
    title = {{Learning and role-playing games}},
    year = {2018},
    booktitle = {Role-playing game studies},
    author = {Hammer, Jessica and To, Alexandra and Schrier, Karen and Bowman, Sarah Lynne and Kaufman, Geoff},
    pages = {283--299},
    publisher = {Routledge}
}

@book{walton1993mimesis,
    title = {{Mimesis as make-believe: On the foundations of the representational arts}},
    year = {1993},
    author = {Walton, Kendall L},
    publisher = {Harvard University Press}
}

@incollection{bowman2018psychology,
    title = {{Psychology and role-playing games}},
    year = {2018},
    booktitle = {Role-playing game studies},
    author = {Bowman, Sarah Lynne and Lieberoth, Andreas},
    pages = {245--264},
    publisher = {Routledge}
}

@article{blatner2009role,
    title = {{Role playing in education}},
    year = {2009},
    journal = {Disponibile all'indirizzo: http://www. blatner. com/adam/pdntbk/rlplayedu. htm},
    author = {Blatner, Adam}
}

@article{winardy2023role,
    title = {{Role, play, and games: Comparison between role-playing games and role-play in education}},
    year = {2023},
    journal = {Social Sciences {\&} Humanities Open},
    author = {Winardy, Gary Collins Brata and Septiana, Eva},
    number = {1},
    pages = {100527},
    volume = {8},
    publisher = {Elsevier}
}

@article{chiu2017role,
    title = {{Role-playing game based assessment to fractional concept in second grade mathematics}},
    year = {2017},
    journal = {Eurasia Journal of Mathematics, Science and Technology Education},
    author = {Chiu, Fu-Yuan and Hsieh, Mei-Ling},
    number = {4},
    pages = {1075--1083},
    volume = {13},
    publisher = {Modestum}
}

@incollection{breazeal2016social,
    title = {{Social robotics}},
    year = {2016},
    booktitle = {Springer handbook of robotics},
    author = {Breazeal, Cynthia and Dautenhahn, Kerstin and Kanda, Takayuki},
    pages = {1935--1972},
    publisher = {Springer}
}

@article{belpaeme2018social,
    title = {{Social robots for education: A review}},
    year = {2018},
    journal = {Science robotics},
    author = {Belpaeme, Tony and Kennedy, James and Ramachandran, Aditi and Scassellati, Brian and Tanaka, Fumihide},
    number = {21},
    volume = {3},
    publisher = {Science Robotics}
}

@article{robaczewski2021socially,
    title = {{Socially assistive robots: The specific case of the NAO}},
    shorttitle = {Socially {\{}{\{}Assistive Robots{\}}{\}}},
    year = {2021},
    journal = {International Journal of Social Robotics},
    author = {Robaczewski, Adam and Bouchard, Julie and Bouchard, Kevin and Gaboury, Sébastien},
    number = {4},
    month = {7},
    pages = {795--831},
    volume = {13},
    doi = {10.1007/s12369-020-00664-7},
    issn = {1875-4805}
}

@book{bloom1956taxonomy,
    title = {{Taxonomy of educational objectives: The classification of educational goals. Handbook 1: Cognitive domain}},
    year = {1956},
    author = {Bloom, Benjamin S and Engelhart, Max D and Furst, Edward J and Hill, Walker H and Krathwohl, David R and {others}},
    publisher = {Longman New York}
}

@article{tarng2010design,
    title = {{The design and analysis of learning effects for a game-based learning system}},
    year = {2010},
    journal = {International Journal of Educational and Pedagogical Sciences},
    author = {Tarng, Wernhuar and Tsai, Weichian},
    number = {1},
    pages = {14--23},
    volume = {4},
    publisher = {Citeseer}
}

@article{ranciere1991ignorant,
    title = {{The ignorant schoolmaster: Five lessons in intellectual emancipation}},
    year = {1991},
    author = {Ranci{\`{e}}re, Jacques}
}

@article{daniau2016transformative,
    title = {{The transformative potential of role-playing games—: From play skills to human skills}},
    year = {2016},
    journal = {Simulation {\textbackslash}{\&} Gaming},
    author = {Daniau, Stéphane},
    number = {4},
    pages = {423--444},
    volume = {47},
    publisher = {Sage Publications Sage CA: Los Angeles, CA}
}

@book{boal2000theater,
    title = {{Theater of the Oppressed}},
    year = {2000},
    author = {Boal, Augusto},
    publisher = {Pluto press}
}

@inproceedings{sanoubari2022robot-mediated,
    title = {{Using Robot-Mediated Applied Drama to Foster Anti-Bullying Peer Support}},
    year = {2022},
    booktitle = {Robophilosophy},
    author = {Sanoubari, Elaheh and Johnson, Amanda and Munoz, John Edison and Houston, Andrew and Dautenhahn, Kerstin}
}

@inproceedings{sanoubari2024makes,
    title = {{What Makes an Educational Robot Game Fun? Framework Analysis of Children’s Design Ideas}},
    year = {2024},
    booktitle = {International Conference on Social Robotics},
    author = {Sanoubari, Elaheh and Mu{\~{n}}oz, John Edison and Yamini, Ali and Randall, Neil and Dautenhahn, Kerstin},
    pages = {40--55},
    organization = {Springer}
}

\end{document}